\documentclass[runningheads]{llncs}
\usepackage{graphicx}

\usepackage{algorithm}
\usepackage{algpseudocode}
\usepackage{amsmath}
\usepackage[misc]{ifsym}
\usepackage{booktabs}

\usepackage{url}

\begin{document}
\title{Training Quantized Deep Neural Networks via Cooperative Coevolution}
%
%
\author{Fu Peng \and
	Shengcai Liu (\Letter)\and
	Ning Lu\and
	Ke Tang}
\authorrunning{ }
%
\institute{Guangdong Key Laboratory of Brain-Inspired Intelligent Computation,\\
	Department of Computer Science and Engineering,\\
	Southern University of Science and Technology, Shenzhen 518055, China 
\email{11930584@mail.sustech.edu.cn,
liusc3@sustech.edu.cn,
11610310@mail.sustech.edu.cn,
tangk3@sustech.edu.cn.}}

\maketitle
\begin{abstract}

This work considers a challenging Deep Neural Network (DNN) quantization task that seeks to train quantized DNNs without involving \textit{any} full-precision operations. Most previous quantization approaches are not applicable to this task since they rely on full-precision gradients to update network weights. To fill this gap, in this work we advocate using Evolutionary Algorithms (EAs) to search for the optimal low-bits weights of DNNs. To efficiently solve the induced large-scale discrete problem, we propose a novel EA based on cooperative coevolution that repeatedly groups the network weights  based on the confidence in their values and focuses on optimizing the ones with the least confidence. To the best of our knowledge, this is the first work that applies EAs to train quantized DNNs. Experiments show that our approach surpasses previous quantization approaches and can train a 4-bit ResNet-20 on the Cifar-10 dataset with the same test accuracy as its full-precision counterpart.

\end{abstract}
\section{Introduction}
Deep Neural Networks (DNNs) are powerful and have a wide range of applications in several fields such as image recognition~\cite{1-he2016deep}, object detection~\cite{2-ren2015faster}, visual segmentation~\cite{3-he2017mask}, text classification~\cite{LiuLCT2022}, etc.
However, DNNs generally require a lot of computational resources.
For example, the size of the well-known VGG-16 model built by Caffe is over 500MB and it consumes 16 GFLOPs, which makes it impractical to be deployed on low-end devices.
Hence, over the past few years, many methods have been proposed to reduce the computational complexity of DNNs, such as pruning~\cite{5-han2015deep}, low-rank decomposition~\cite{6-yu2017compressing}, knowledge distillation~\cite{7-xu2019positive}, and quantization~\cite{8-courbariaux2015binaryconnect,5-han2015deep,14-hubara2016binarized,15-rastegari2016xnor,9-wu2018training,19-DBLP:conf/cvpr/WuWGLCLHLY20,17-DBLP:conf/icml/ZhaoHDSZ19,12-zhou2016dorefa}. 
Specifically, DNN quantization maps the network weights from high bits to low bits, significantly reducing memory usage, speeding up inference, and enabling the deployment of networks on mobile devices with dedicated chips~\cite{google-whitepaper}.

Although existing quantization approaches have achieved notable success, most of them rely on full-precision gradients to update the network weights~\cite{8-courbariaux2015binaryconnect,14-hubara2016binarized,15-rastegari2016xnor,12-zhou2016dorefa}, hindering their practical usages. In real-world applications, one may need to quantize a pre-trained full-precision DNN on different low-end devices for better adaptability, and the quantization procedure that is conducted on the device cannot involve any full-precision operations~\cite{9-wu2018training}.
On the other hand, as a powerful search framework, EAs do not use any gradient information \cite{LiuWTQY15,LIUTY2021}, which is naturally suitable for this scenario.
Therefore, in this work we advocate using EAs to search for the low-bits weights of quantized DNNs.

Specifically, we first formulate DNN quantization as a large-scale discrete optimization problem. 
Since this problem involves a huge number of variables (network weights), e.g., ResNet-20 has 269722 parameters, we propose a novel EA based on cooperative coevolution to solve it. 
Given a pre-trained full-precision DNN, our algorithm first quantizes it to obtain an initial solution and then leverages estimation of distribution algorithm (EDA) to optimize the low-bits weights.
To improve search efficiency, the algorithm repeatedly groups the network weights according to the confidence in their values and focuses on optimizing the ones with the least confidence. 
Finally, we compare our algorithm with exiting quantization approaches by applying them to train a 4-bit ResNet-20 on the Cifar-10 dataset, without involving any full-precision operations. 
The results show that our algorithm performs better and the quantized DNN obtains the same test accuracy as its full-precision counterpart, i.e., quantization without loss of accuracy.
In summary, we make the following contributions in this paper:
\begin{enumerate}
    \item We propose a novel EA based on cooperative coevolution to train quantized DNNs. To the best of our knowledge, this is the first work that applies EAs to search for the optimal low-bits weights of DNNs.
    \item We conduct experiments to verify the effectiveness of the proposed algorithm. Notably, it can train a 4-bit ResNet-20 without accuracy degradation compared to the full-precision DNN, which indicates the great potential of EAs in DNN quantization.
\end{enumerate}

\section{Related Work}
This section presents a brief literature review on the field of DNN quantization and cooperative coevolution.

\subsection{DNN Quantization}
DNN quantization is a popular research area, and researchers have proposed many quantization approaches~\cite{google-whitepaper}, which can be classified into two categories: quantization-aware training (QAT) and post-training quantization (PTQ).

PTQ directly quantizes well-trained full-precision networks without re-training\\\cite{19-DBLP:conf/cvpr/WuWGLCLHLY20}.
Two representative PTQ approaches are Outlier Channel Splitting (OCS)~\cite{17-DBLP:conf/icml/ZhaoHDSZ19} and Deep Compression~\cite{5-han2015deep}.
The former deals with outliers during quantization by duplicating channels containing outliers and halving the channel values. The latter introduces a three-stage pipeline: pruning, trained quantization, and Huffman coding, which work together to reduce the memory storage for DNNs.

Unlike PTQ, QAT quantizes and finetunes network parameters in the training process~\cite{242-Loss-Aware}, which usually obtains better performance, thus attracting much more research interest.
BinaryConnect~\cite{8-courbariaux2015binaryconnect} restricts the weights to two possible values, i.e., -1 or 1, but the activations are still full-precision.
BNN~\cite{14-hubara2016binarized} quantizes both weights and activations to -1 or 1.
XNOR-Net~\cite{15-rastegari2016xnor} proposes filter-wise scaling factors for weights and activations to minimize the quantization error. To further accelerate the training of DNNs, some work also attempts to quantize gradients. DoReFa-Net~\cite{12-zhou2016dorefa} uses quantized gradients in the backward propagation, but the weights and gradients are stored with full precision when updating the weights as the same as previous works.
To the best of our knowledge, WAGE~\cite{9-wu2018training} is currently the only work that updates the quantized weights with discrete gradients.

\subsection{Cooperative Coevolution}
Cooperative coevolution is a powerful framework that leverages the “divide-and-conquer” idea to solve large-scale optimization problems.
As first shown by Yang and Tang~\cite{28-DBLP:conf/ppsn/ChenWYT10,Tang2-DBLP:conf/ideal/LiuT13,13-DBLP:journals/tec/MaLZTLXZ19,13-yang2008large,Tang3-DBLP:conf/cec/YangTY08a}, the framework of cooperative coevolution consists of three parts: problem decomposition, subcomponent optimization, and subcomponents coadaptation.
Among them, problem decomposition is the key step~\cite{18-potter2000cooperative}. An effective decomposition can ease the optimization difficulty of a large-scale problem~\cite{20-DBLP:journals/tec/SunKH18}. In contrast, an improper decomposition may lead the algorithms to local optimums~\cite{ma2022merged,13-DBLP:journals/tec/MaLZTLXZ19,21-DBLP:journals/tec/SonB04}. 

There are three categories of problem decomposition approaches: static decomposition, random decomposition, and learning-based decomposition\cite{ma2022merged}. Static decomposition approaches do not take account into the subcomponents interactions and fixedly decompose the decision variables into subcomponents~\cite{22-DBLP:journals/tec/vandenBerghE04,23-DBLP:conf/cec/CaoWS0RJL15,12-DBLP:conf/ppsn/PotterJ94}. Conversely, selecting decision variables randomly for each subcomponent is the main idea of random decomposition approaches\cite{26-DBLP:journals/tec/LiY12,25-DBLP:conf/cec/OmidvarLYY10,27-DBLP:journals/isci/TrunfioTW16,13-yang2008large}. One of the most famous random decomposition methods is EACC-G proposed by Yang and Tang~\cite{13-yang2008large}. The main idea of this method is to divide the interdependent variables into the same subcomponent, but the dependencies between subcomponents should be as weak as possible. The learning-based approaches try to discover the interactions between variables~\cite{28-DBLP:conf/ppsn/ChenWYT10,8-DBLP:journals/toms/MeiOLY16,14-DBLP:journals/tec/OmidvarLMY14}.

\section{Method}

In this section, we first formulate DNN quantization as a large-scale discrete optimization problem and introduce our quantization functions. Then we detail the EDA applied to this problem. Finally, to further improve the algorithm performance, the cooperative coevolution framework is proposed. 

\subsection{Problem Formulation}
In a DNN with $L$ layers, let $\boldsymbol{w_l}$ represent the full-precision weights and $\boldsymbol{\hat{w_l}}$ represent the $k$ bits quantized weights at layer $l$, which both are an $n_l$-dimension vector, i.e., there are $n_l$ paremeters at layer $l$. We combine the quantized weights from all layers into $\boldsymbol{\hat{w}}=[\boldsymbol{\hat{w}_1},\boldsymbol{\hat{w}_2},…,\boldsymbol{\hat{w}_L}]$. The parameters in $\boldsymbol{\hat{w}}_l$ can only take one of $2^k$ possible discrete values, i.e., $\boldsymbol{\hat{w}_l} \in \{t_{l1}, t_{l2}, …, t_{l2^{k}}\}^{n_l}$. We formulate the DNN low-bit quantization problem as the following large-scale discrete optimization problem.

\begin{equation}
\max\limits_{\boldsymbol{\hat{w}}} f(\boldsymbol{\hat{w}})\quad s.t.\quad  \boldsymbol{\hat{w}}_l\in \{t_{l1}, t_{l2}, …, t_{l2^{k}}\}^{n_l}, \quad l = 1,2,...,L,
\label{equ:optimization_problem}
\end{equation}
where $f(\boldsymbol{\hat{w}})$ represents the accuracy of the quantized DNN. Since a DNN usually has a huge number of parameters, e.g., ResNet-152 has around 11 million paremeters, this is a large-scale discrete optimization problem.  

\subsection{Quantization Functions}
To obtain a quantized DNN and construct the search space of our algorithm, we need to identify all the possible discrete values for each weight and activation. Moreover, the initial solution of our algorithm is obtained from a full-precision DNN. Based on the above two considerations, we design two linear quantization functions to map full-precision weights and activations to discrete ones separately.

For the weights, let the maximum and minimum values of each layer weights $\boldsymbol{w_l}$ be $[w_l^{min}, w_l^{max}]$, then the full-precision weights $\boldsymbol{w_l}$ at layer $l$ are discretized with a uniform distance $\delta_l$:

\begin{equation}
\delta_l(k)=\frac{w_l^{max}-w_l^{min}}{2^k-1},
\end{equation}
where $k$ is the number of bits. The quantization function for weights can be represented as:

\begin{equation}
Q(\boldsymbol{w_l}) = Clip\{round(\frac{\boldsymbol{w_l}}{\delta_l(k)}) \cdot \delta_l(k), w_l^{min}, w_l^{max}\},
\label{eq:weight_quantization}
\end{equation}
where the $Clip$ function is the saturation function, and the $round$ function maps continuous values to their nearest integers. 

For the remaining parameters including activations and the parameters in batch normalization layers, we assume that the range of parameters is $[-1,1]$ as WAGE~\cite{9-wu2018training}. The quantization function for activations $\boldsymbol{a_l}$ at layer $l$ can be represented as:
\begin{equation}
Q(\boldsymbol{a_l}) = round(\frac{\boldsymbol{a_l}}{\delta_l(k)}) \cdot \delta_l(k).
\label{eq:activation_quantization}
\end{equation}

\subsection{Estimation of Distribution Algorithm}
We propose to use the Estimation of Distribution Algorithm (EDA) to search for discrete weights.
The overall framework of EDA for training quantized DNNs is summarized in Algorithm \ref{alg::EDA}.
We encode the quantized DNN weights into a fixed-length 1-dimensional array as the representation of our solution, i.e., $\boldsymbol{\hat{w}}=[\hat{w}_1,\hat{w}_2, …, \hat{w}_n]$, where $n$ represents the total number of parameters in a DNN. Then we construct a probabilistic model over it. For simplicity, we assume that the weights of the neural network are all independent of each other like PBIL~\cite{22-baluja1994population}. Specifically, for each weight $\hat{w}_i$, there are $2^k$ possible values. Each possible value corresponds to a probability $p_j$, where $j =1, 2, …,2^k $, $k$ is the bit length of weights, and $\sum_{j=1}^{2^k}{p_j}=1$. After the initial quantized network is obtained (Line 1), i.e., $\boldsymbol{\hat{w}}=[\hat{w}_1=a_1,\hat{w}_2=a_2, …, \hat{w}_n=a_n]$, we initialize the probabilistic model $P$ of the weights using $\sigma$-greedy strategy (Line 2), which is shown as Eq. (\ref{eq:sigma-greedy}):
\begin{equation}
\left\{
    \begin{array}{lc}
        P(\hat{w}_i=a_i) = \sigma, & \\
        P(\hat{w}_i= \text{one of the other possible values}) = \displaystyle{\frac{1-\sigma}{2^k-1}}.
    \end{array}
\right.
\label{eq:sigma-greedy}
\end{equation}
 That is, if $\hat{w}_i$ takes the value $a_i$, then $P(\hat{w}_i=a_i) = \sigma$. The probability of other possible values is $(1-\sigma)/(2^k-1)$, where $0<\sigma<1$. For each generation, we sample weights from the probabilistic model to generate new individuals (Line 4), get the fitness values of them (Line 5) and rank them by their fitness values in descending order (Line 6). To update the probabilistic model $P$, we calculate the probability of each possible value for $w_i$ according to the first $N_{best}$ new individuals and construct the probabilistic model $P_{best}$ of them (Line 9). Finally, we update $P$ using $(1-\alpha)P+\alpha P_{best}$ (Line 10), where $\alpha$ is updating step.
\begin{algorithm}[t]
	\caption{Estimation of Distribution Algorithm} 
	\label{alg::EDA}
	\begin{algorithmic}[1]
		\Require
		the number of best individuals $N_{best}$ to update the probabilistic model;
		updating step $\alpha$ ; 
		generation number $G$;
		the size of population $S$
		\Ensure
		best individual $I_{best}$
		\State Initialize best individual $I_{best}=[\hat{w}_1=a_1,\hat{w}_2=a_2, …, \hat{w}_n=a_n]$
		\State Initialize probabilistic model $P$ using $\sigma$-greedy strategy
		\For{generation $i$ from 0 to $G$} %
		\State Generate $S$ new individuals according to  $P$
		\State Get the fitness values of the new individuals 
		\State Rank the new individuals by fitness values in descending order
		\State Update the best individual $I_{best}$
		\State Select the first $N_{best}$ best individuals
		\State Construct the probabilistic model $P_{best}$ of the $N_{best}$ best individuals
		\State Update the probabilistic model $P$: $P = (1-\alpha)P+\alpha P_{best}$
		\EndFor
	\end{algorithmic}
\end{algorithm}

\subsection{Cooperative Coevolution}
Since our optimization problem has a huge number of decision valuables, to further improve the search efficiency, we propose a novel cooperative coevolution algorithm based on EDA inspired by Yang and Tang~\cite{13-yang2008large}, namely EDA+CC.

\begin{figure}[t]
	\centering
	\includegraphics[scale=0.3]{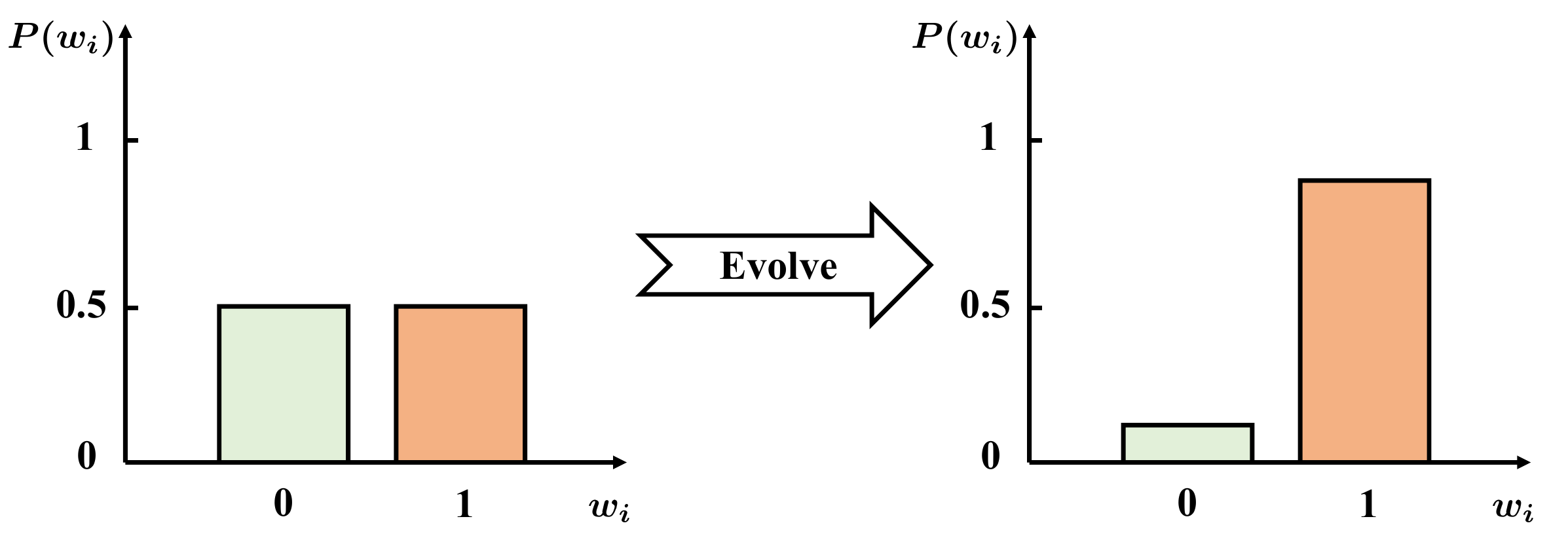}
	\caption{$P(w_i)$ converges gradually as EDA evolving}
	\label{fig::EDA}
\end{figure}

The most important part of the cooperative coevolution algorithm lies in the efficient grouping of variables. As the EDA searches, the probabilistic model $P$ gradually converges. However, different decision variables have different convergence rates. Fig.~\ref{fig::EDA} gives a simple example of the convergence of the probabilistic model when applying EDA to a $0/1$ optimization problem. Suppose the decision variables are encoded as $\boldsymbol{w}=[w_1,w_2,…,w_n]$ and variables are independent of each other. Initially, $P(w_i=1) = 0.5$. As the evolution proceeds, $P(w_i=1)$ gradually converges to $1$. For $w_i$, if $P(w_i)$ converges quickly, it intuitively shows that EDA is confident about the value of $w_i$, which means $w_i$ should not be changed in the subsequent searching process; conversely, if $P(w_i)$ converges slowly, then $w_i$ should be further optimized.

\begin{algorithm}[t]
	\caption{EDA+CC} 
	\label{alg::EDA+CC}
	\begin{algorithmic}[1]
		\Require
		the number of best individuals $N_{best}$ to update the probabilistic model;
		updating step $\alpha$ ; 
		generation number $G$;
		the size of population $S$;
		accuracy threshold $T$;
		flag $F$;
		total number of weights $n$;
		random number $\beta$
		\Ensure
		best individual $I_{best}$
		\State Initialize best individual $I_{best}=[\hat{w}_1=a_1,\hat{w}_2=a_2, …, \hat{w}_n=a_n]$
		\State Get the fitness values of $I_{best}$
		\State Initialize probabilistic model $P$ using $\sigma$-greedy strategy
		\State $F \leftarrow 0$
		\While{fitness value of $I_{best}$ $\leq$ $T$ }
		\If{F=1}
		\State Sort the weights by the convergence speed of $P$ in descending order
		\State Divide the first $\beta \cdot n$ weights into group A
		\State Divide the remaining weights into group B  
		\State Reinitialize the probabilistic model over group B using $\sigma$-greedy strategy
		\EndIf
		\For{generation $i$ from 0 to $G$} %
		\State Generate $S$ new individuals according to  $P$
		\State Get the fitness values of the new individuals 
		\State Rank the new individuals by fitness values in descending order
		\State Update the best individual $I_{best}$
		\State Select the first $N_{best}$ best individuals
		\State Construct the probabilistic model $P_{best}$ of the $N_{best}$ best individuals
		\State Update the probabilistic model $P$: $P = (1-\alpha)P+\alpha P_{best}$
		\State $F \leftarrow 1$
		\EndFor
		\EndWhile
		
	\end{algorithmic}
\end{algorithm}

\begin{figure}[t]
	\centering
	\includegraphics[width=\textwidth]{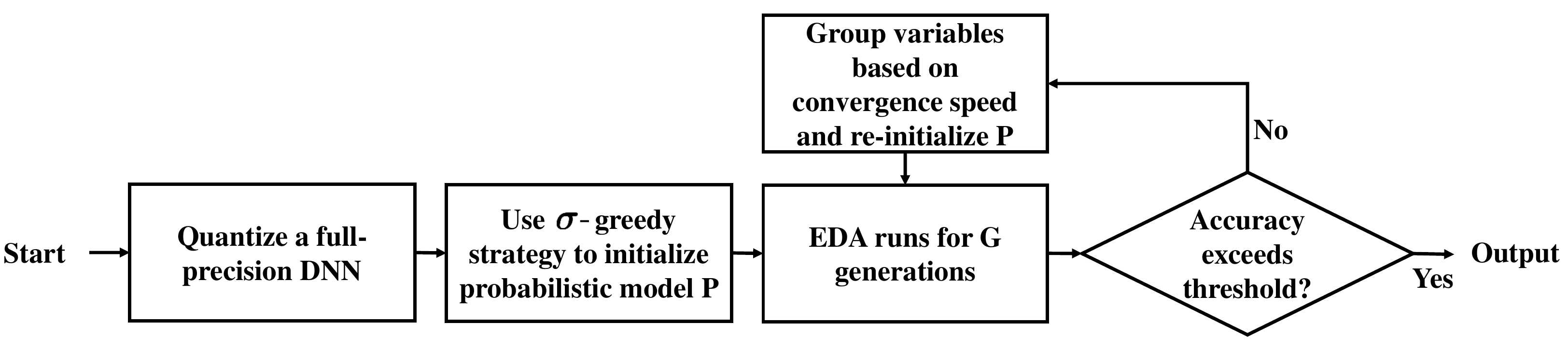}
	\caption{EDA+CC framework} 
	\label{fig::EDA+CC-framwork}
\end{figure}

Based on this intuition, we group the decision variables according to the confidence in their values, i.e., the speed of convergence. Specifically, we rank the decision variables according to the convergence speed of the probabilistic model in descending order during the EDA run. We divide the first $\beta \cdot n$ variables (which converge fast) into one group and the remaining variables (which converge slowly) into another group, where $\beta \in [0,1]$ is a random number and $n$ is the total number of weights in the network. For the former, we fix them. For the latter, we first perturb the probabilistic model of them with the $\sigma$-greedy strategy and then use EDA to optimize them. Fig.~\ref{fig::EDA+CC-framwork} shows the framework of EDA+CC. Detials of EDA+CC are shown in Algorithm \ref{alg::EDA+CC}. Every $G$ generations we regroup the variables, perturb the probabilistic model, and run EDA again until the network accuracy reaches the threshold.

\section{Experiments}
We use EDA+CC to train 4-bit quantized ResNet-20 on the Cifar-10 training set and test its performance on the test set. 
Firstly, we compare our EDA+CC algorithm with WAGE~\cite{9-wu2018training} (See Section~\ref{section: 4.2}).  WAGE is the only work that can update discrete weights with quantized gradients.  
Secondly, we investigate the influence of different initial quantized DNNs by ablation study (See Section~\ref{section 4.3}).
Finally, besides EDA we also test the performance of Genetic Algorithm (GA)~\cite{20-mitchell1991royal} and Local Search Algorithm (LS)~\cite{21-hoos2004stochastic} (See Section~\ref{section 4.4}).

\subsection{Experiment Settings} \label{section: 4.1}
We implement EDA+CC based on TensorFlow 2.1.0 with python 3.6.9 and run the experiments on Nvidia RTX 2080ti. The settings of algorithms are as follows. The number of generations $G$ is 500, the size of population $S$ is 20, the number of best individuals $N_{best}$ for each generation is 20, the updating step $\alpha$ is 0.1, and the parameter $\sigma$ in the $\sigma$-greedy strategy is 0.95. To enforce the randomness of the algorithm, we set $\beta$ as a random variable that obeys uniform distribution, $\beta \sim N(0.4,0.6)$.

In the following, we show how to construct an initial quantized network, in which the range of weights of each layer is known. First, we use Eq. (\ref{eq:weight_quantization}) and (\ref{eq:activation_quantization}) to quantize the pre-trained full-precision ResNet-20. To increase the randomness of the initial quantized network we randomly select s\% of all its parameters and perturb them to adjacent values ($s=20,30,40$ separately). We denote the quantized network obtained by the above process as ResNetQ-Switch-s\% and the pre-trained full-precision network as ResNet20-Float. Table \ref{table::accuracy-of-diff-inital-network} summarizes the accuracies of the different initial quantized DNNs. 

Since ResNet-20 has 269722 parameters, to reduce the search space, we restrict weights to two possible values, i.e., the value before perturbation and the value next to it. Thus, the problem becomes a binary optimization problem and the search space size is $2^{269722}$. Note that this is still a huge search space, it is $10^{81022}$ times larger than the Go search space.
 
\begin{table}[t]
	\centering
	\caption{Accuracies of different initial quantized networks obtained by perturbing different proportional parameters. We also list the accuracy of the pre-trained full-precision network, namely ResNet20-Float.}
	\setlength{\tabcolsep}{5mm}
	\begin{tabular}{ccc}
        \toprule
		Network
		&\begin{tabular}[c]{@{}c@{}} Training Set\end{tabular}   
		&\begin{tabular}[c]{@{}c@{}} Test Set\end{tabular} \\ 
		\midrule
		ResNet20-Float& 98.65\%&91.00\%\\
		ResNet20Q-Switch-20\%& 88.20\%&85.38\%\\
		ResNet20Q-Switch-30\%&46.65\%&46.79\%\\
		ResNet20Q-Switch-40\%&20.32\%&19.60\%\\
		\bottomrule
	\end{tabular}
	\label{table::accuracy-of-diff-inital-network}
\end{table}

\subsection{Verifying the Effectiveness of EDA+CC}\label{section: 4.2}

To verify the effectiveness of EDA+CC, we first compare EDA+CC with WAGE\\~\cite{9-wu2018training}, a representative quantization method that quantizes gradients. 
The code of WAGE is available at~\cite{wage-github}. 
To further examine the performance of the cooperative coevolution algorithm, we also compare EDA+CC with EDA w/o CC. EDA w/o CC re-initializes the probabilistic model using the $\sigma$-greedy strategy without grouping the decision valuables when EDA restarts. Both EDA+CC and EDA w/o CC use 150K fitness evaluations and take about 23.3 hours separately, in which the time complexity is acceptable. 
 
\begin{figure}[t]
	\centering
	\includegraphics[scale=0.2]{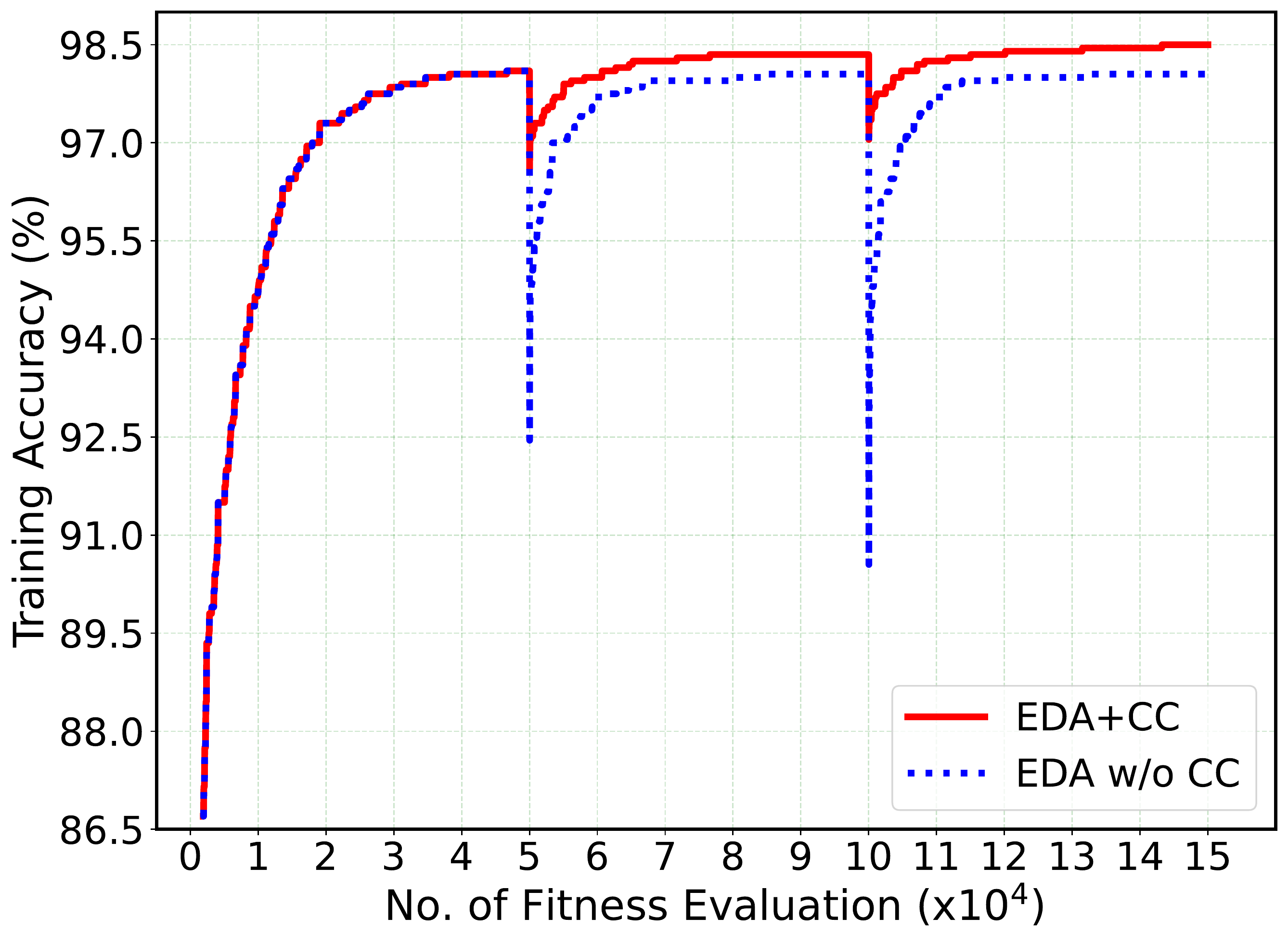}
	\caption{Training curves of EDA+CC and EDA w/o CC.} 
	\label{fig::training_curve_for_EDA+CC}
\end{figure}

Table \ref{table::Compare_EDA+CC} shows the accuracies of the quantized DNNs obtained by different approaches. We use ResNet20Q-Switch-30\% as our initial solution. The initial accuracy of ResNet20Q-Switch-30\% is 46.50\%. For EDA+CC, the training set accuracy only decreases by 0.15\% and the test set accuracy increases by 0.4\% compared to the full-precision network. In comparison, the accuracy of the network obtained by WAGE training is only about 43\%, which is much worse than EDA+CC. We speculate the reason for the poor performance of WAGE might be that WAGE is designed for quantized DNNs with 2-bit weights and 8-bit activations, while 
our paper uses a more rigorous and hardware-friendly setting: 4-bit weights and 4-bit activations quantized DNNs. Comparing EDA+CC with EDA w/o CC, we can see the positive effect of cooperative coevolution. Applying cooperative coevolution increases the training accuracy from 98.05\% to 98.50\% and the testing accuracy from 89.40\% to 91.40\%. The effectiveness of cooperative coevolution is mainly shown in two aspects: improving the quality of the solution and accelerating the convergence. Fig.~\ref{fig::training_curve_for_EDA+CC} shows the training curves of EDA+CC and EDA w/o CC, i.e., the accuracy of the best individual in each generation. As Fig.~\ref{fig::training_curve_for_EDA+CC} shows, after using the $\sigma$-greedy strategy to re-initialize the probabilistic model $P$ and restarting EDA, EDA+CC can accelerate the convergence and help EDA find a better solution.

\begin{table}[t]
	\caption{Compare EDA+CC with WGAE and EDA w/o CC.}
	\setlength{\tabcolsep}{5mm}
	\centering
	\begin{tabular}{ccc}
		\toprule
		Algorithm& Training Set & Test Set \\ \midrule
		EDA+CC& \textbf{98.50\%} & \textbf{91.40\%} \\
		EDA w/o CC&98.05\%&89.40\%  \\
		WAGE&43.44\%&41.35\% \\ \bottomrule
	\end{tabular}

	\label{table::Compare_EDA+CC}
\end{table}

\begin{table}[t]
	\caption{Results of different initial quantized networks.}
	\setlength{\tabcolsep}{5mm}
	\centering
	\begin{tabular}{ccccc}
		\toprule
		Initial Quantized DNN& Training Set & Test Set &No. of FEs \\ \midrule
		ResNet20Q-Switch-20 & 99.25\%  & 91.50\%& 50K\\ 
		ResNet20Q-Switch-30 & 98.50\%  & 91.40\%& 150K\\ 
		ResNet20Q-Switch-40& 90.09\%  & 82.75\%& 150K \\ \bottomrule
	\end{tabular}
	\label{table::result-diff-initial-network}
\end{table}

\subsection{Ablation Study}\label{section 4.3}
We conduct more detailed studies on different initial quantized networks for EDA+CC. Table~\ref{table::result-diff-initial-network} shows the accuracies of the quantized networks obtained by EDA+CC with different initial networks. It can be seen that EDA+CC reaches 90.09\% accuracy after 150K fitness evaluations for ResNet20Q-Switch-40. We estimate that it will take about 500K fitness evaluations(FEs)  for EDA+CC to reach around 98\% accuracy because each restart of EDA with $\sigma$-greedy strategy can improve the accuracy by about 0.9\%. In summary, Table~\ref{table::result-diff-initial-network} illustrates that as the initial accuracy decreases, EDA+CC requires more FEs to train a quantized DNN without accuracy decay compared to the full-precision network.

\begin{figure}[t]
	\centering
	\includegraphics[scale=0.2]{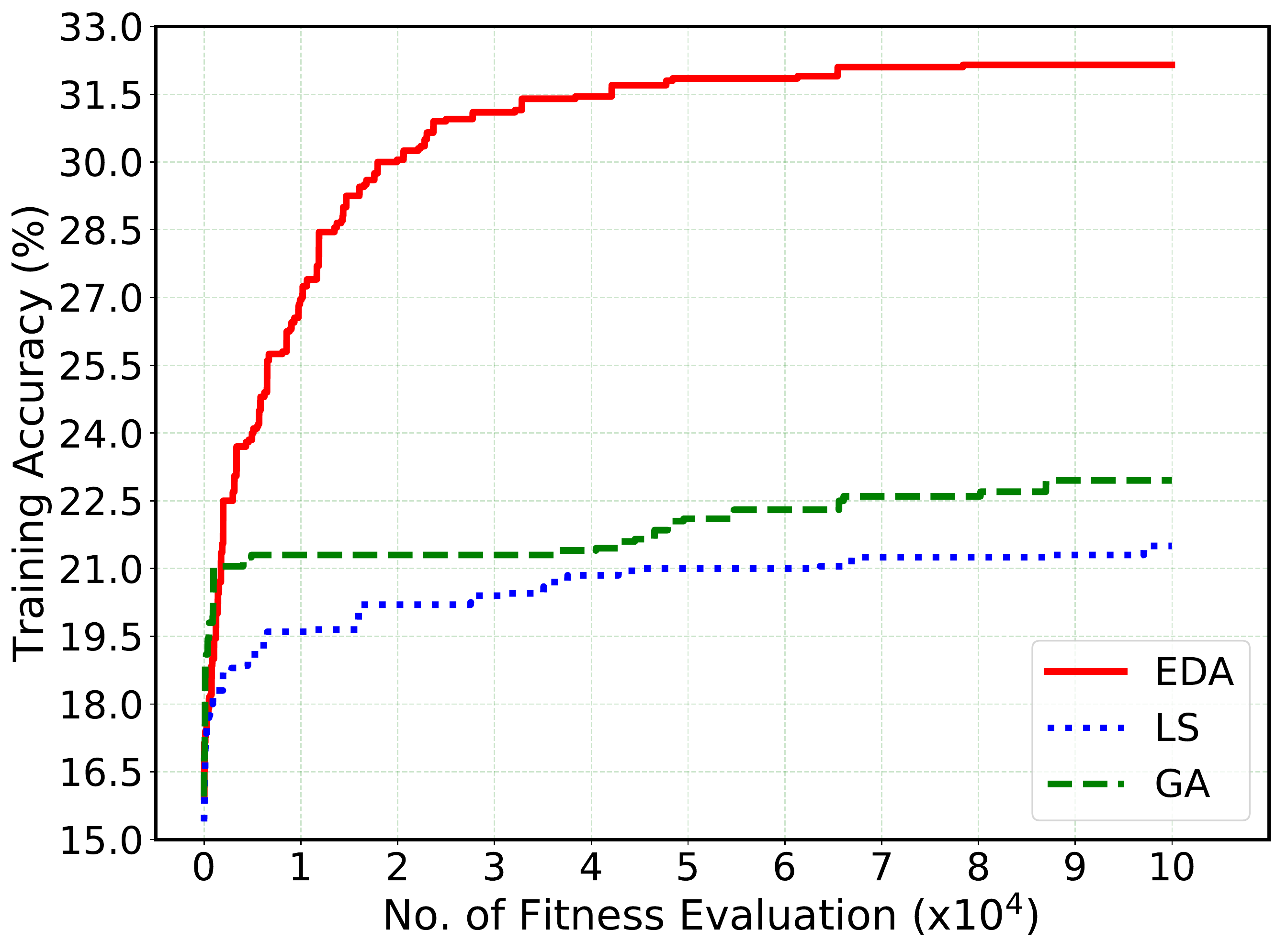}
	\caption{Training curves of EDA, GA and LS.} 
	\label{fig::training_curve_for_3_algo}
\end{figure}

\subsection{Comparison of EDA, GA and LS} \label{section 4.4}
We compare three search algorithms, GA, LS, and EDA. We use ResNet20Q-Switch-50\% as the initial quantized network. Each algorithm uses 100K fitness evaluations. Fig.~\ref{fig::training_curve_for_3_algo} shows the training curves of the three algorithms. It can be seen that EDA performs significantly better than LS and GA, which indicates that the distribution estimation mechanism is more suitable than the crossover and mutation mechanisms for the problem considered in this study. The crossover and mutation mechanisms might break some good patterns in the individuals imperceptibly, while the distribution estimation mechanism optimizes the individuals in a global way. It is worth noting that, theoretically, in the binary space, ResNet20Q-Switch-50\% corresponds to random initialization, because half of the parameters are randomly perturbed. All three algorithms can obtain better accuracy than ResNet20Q-Switch-50\%, which illustrates the potential of search-based algorithms in training quantized DNNs.

\section{Conclusion and Future Work}
In this paper, we investigate search-based training approaches for quantized DNNs, focusing on exploring the application of cooperative coevolution to this problem. Unlike existing works, EDA+CC does not need gradient information. Considering the search space of this problem is extremely large (e.g., in our experiments it is $10^{81022}$ times larger than the Go search space), we propose to use cooperative coevolution to help solve this problem.
The results show that our method can obtain quantized networks without accuracy decay compared to floating-point networks in our experiment setting.

Overall, this work is a proof of concept that EAs can be applied to train quantized DNNs. There are many subsequent lines of research to pursue, e.g., the effects of other variable grouping mechanisms.
Moreover, the method of determining the ranges of discrete values should also be studied.
Finally, based on the cooperative coevolution framework, it is interesting to investigate on solving different sub-problems by different algorithms \cite{LiuT019,LiuTL020,LiuTY22,TangLYY21}, hopefully leading to better optimization performance.

%
%
\bibliographystyle{splncs04}
\bibliography{quantization_paper}

\end{document}